\documentclass[letterpaper, 10 pt, journal, twoside]{IEEEtran}
\usepackage{ifpdf}

\usepackage{cite}

\ifCLASSINFOpdf
  \usepackage[pdftex]{graphicx}
\else
  \usepackage[dvips]{graphicx}
  \graphicspath{{../eps/}}
  \DeclareGraphicsExtensions{.eps}
\fi
\usepackage{amsmath,amssymb,amscd,amsfonts,amsthm}
\usepackage[hidelinks]{hyperref}
\usepackage[nameinlink]{cleveref}
\Crefname{figure}{Fig.}{Figs.}

\usepackage{colortbl}
\usepackage[dvipsnames]{xcolor}
\definecolor{Gray}{gray}{0.85}
\definecolor{LightCyan}{rgb}{0.88,1,1}

\usepackage{algorithmic}

\usepackage{booktabs}
\usepackage{array}

\usepackage{graphicx} 
\usepackage{subcaption}
\usepackage{caption}
\usepackage{fixltx2e}
\usepackage{dblfloatfix}
\usepackage{microtype}
\usepackage{url}
\usepackage{soul}

\begin{document}
%
\title{MUSE: A Real-Time Multi-Sensor State Estimator for Quadruped Robots}
%
%
%

\author{Ylenia Nisticò$^{1,2,*}$,
        João Carlos Virgolino Soares$^{1}$,
        Lorenzo Amatucci$^{1,2}$,
        Geoff Fink$^{1,3}$,
        and~Claudio Semini$^{1}$
\thanks{Manuscript received: November, 3, 2024; Revised January, 3, 2025; Accepted March, 2, 2025.}
\thanks{This paper was recommended for publication by
Editor Pascal Vasseur upon evaluation of the Associate Editor and Reviewers' comments. \\
This work was supported by the ASI PEGASUS Project.} 
\thanks{$^1$ Dynamic Legged Systems (DLS), Istituto Italiano di Tecnologia (IIT),
Genoa, Italy. \{\texttt{first\_name.first\_surname}\}\texttt{@iit.it}}
\thanks{$^2$ Dipartimento di Informatica, Bioingegneria, Robotica e Ingegneria dei Sistemi (DIBRIS), 
University of Genoa, Genoa, Italy}
\thanks{$^3$ Department of Engineering, Thompson Rivers University, Kamloops, BC, Canada.
 \texttt{gefink@tru.ca}}
\thanks{$^*$Corresponding author: \texttt{ylenia.nistico@iit.it}}
\thanks{Digital Object Identifier (DOI): 10.1109/LRA.2025.3553047}
}

%
%

\markboth{IEEE Robotics and Automation Letters. Preprint Version. Accepted March, 2025}
{Nisticò \MakeLowercase{\textit{et al.}}: MUSE: A Real-Time Multi-Sensor State Estimator for Quadruped Robots} 
%



\maketitle

\begin{abstract} 
This paper introduces an innovative state estimator, MUSE (MUlti-sensor State Estimator), designed to 
enhance state estimation's accuracy and real-time performance in quadruped 
robot navigation. The proposed state estimator builds upon our previous work presented in \cite{fink20iros}. It integrates data from a range of onboard sensors, including IMUs, encoders, cameras, and LiDARs, to deliver a comprehensive and 
reliable estimation of the robot's pose and motion, even in slippery scenarios. 
We tested MUSE on a Unitree Aliengo robot, successfully closing the locomotion control loop in difficult scenarios, including slippery and uneven terrain. Benchmarking against Pronto~\cite{pronto2020camurri} and VILENS~\cite{wisth2022vilens} showed 67.6\% and 26.7\% reductions in translational errors, respectively. Additionally, MUSE outperformed DLIO~\cite{chen23dlio}, a LiDAR-inertial odometry system in rotational errors and frequency, while the 
proprioceptive version of MUSE (P-MUSE) outperformed TSIF~\cite{bloesch2018twostate}, with a 45.9\% reduction in absolute trajectory error (ATE).
\looseness=-1
\end{abstract}


%
\IEEEpeerreviewmaketitle

\section{Introduction}
\label{sec:introduction}
\IEEEPARstart{Q}{uadruped} robots are renowned for their ability to traverse difficult terrains and, for this reason, they have gained increasing importance in fields such as academic research, inspection, and monitoring. Perceptive information is crucial in unstructured environments~\cite{grandia2023perceptive}, where accurate state estimation enables robust locomotion, real-time feedback, and stable gait control.
\looseness=-1

This work introduces \textbf{MUSE}, a \textbf{MU}lti-sensor \textbf{S}tate \textbf{E}stimator for quadruped robots, emphasizing coordinated sensing, real-time data processing, and refined fusion algorithms, critical for dynamic movement control and autonomous task execution.

Early research in state estimation has focused on combining proprioceptive (e.g., IMU, encoders, torque sensors) and exteroceptive (e.g., cameras, LiDARs) sensors. Exteroceptive sensors provide accurate, low-drift pose estimates, essential for autonomous navigation and SLAM~\cite{cadena16slam}, but may fail under adverse conditions or introduce time delays due to limited frequencies. Hence, they are often paired with high-frequency proprioceptive measurements, which remain reliable in settings where exteroceptive sensors struggle (e.g., poor lighting or limited features). Notable sensor-fusion methods include ORB-SLAM3~\cite{campos2021orb3}, a versatile framework for visual-inertial and multi-map SLAM, and DLIO~\cite{chen23dlio}, known for its computationally efficient LiDAR-inertial odometry.

Specifically for legged robots, state-of-the-art methods take into account also leg kinematics, which provides detailed information about the movement and positioning of each leg, and can help to improve the estimate.
Part of the work in legged robots state estimation has focused solely on proprioception.
For instance, the study in~\cite{bloesch2013state} employs an Observability Constrained Extended Kalman filter to estimate foothold positions and overall robot pose without assuming fixed terrain geometry, \cite{bloesch2018twostate} propose TSIF, a residual-based recursive filter for state estimation in dynamic systems without requiring explicit process models, 
while~\cite{hartley2020contact} uses an Invariant Extended Kalman Filter (InEKF), to fuse contact-inertial dynamics with forward kinematic corrections.
\looseness=-1

However, proprioceptive-only approaches often suffer from drift, especially on soft or slippery terrains. A study done in~\cite{fahmi2021state} examining the impact of soft terrain revealed that relying solely on proprioception, with an estimator assuming rigid contacts, led to significant drift compared to navigating rigid terrain. Techniques such as \cite{bloesch2013stateslippery} and \cite{jenelten2019dynamic} address this by filtering out outliers and modeling slip, while \cite{wisth2020preintegrated} uses factor graphs to account for uncertainty in contact points. In another work~\cite{santana2024proprioceptive}, the authors introduced an innovative InEKF designed specifically for legged robots, relying solely on proprioceptive sensors and incorporating robust cost functions in the measurement update. Although the use of these robust cost functions significantly reduced drift, they were not able to completely eliminate it.

These studies investigating the impact of terrains on legged robots' state estimation reveal the limitations of assuming non-slip conditions in the state estimator. While avoiding this assumption can lead to improved results, proprioceptive state estimation alone falls short of providing a drift-free pose, emphasizing the necessity of incorporating exteroceptive sensor data for enhanced accuracy.

To achieve low-drift estimates, many works fuse leg kinematics and inertial measurements with exteroceptive sensors. An example of a multi-sensor state estimator is Pronto~\cite{pronto2020camurri}, which integrates stereo vision and LiDAR data into an EKF that also combines IMU and leg kinematics.
Although effective, Pronto is built with a non-slip assumption, meaning it operates under the premise that the robot maintains constant contact with the ground without experiencing slippage or falling. This assumption simplifies the state estimation process but overlooks the possibility of slippage or loss of contact, which can significantly impact the accuracy and reliability of the estimated state, particularly in challenging terrains.
In~\cite{teng2021legged} an InEKF is employed for state estimation in a bipedal robot navigating slippery terrain. This method fuses Realsense T265 vision with inertial and leg-kinematic data, using an online noise parameter to adapt to measurement noise.
STEP~\cite{kim2022step}, instead,  adopts a stereo camera for speed estimation and pre-integrated foot velocity factors, bypassing explicit contact detection and non-slip assumption.
However, it is worth noting that these state estimators heavily depend on camera inputs, which may sometimes be unreliable, potentially affecting the accuracy and robustness of the estimated states.
VILENS, proposed in~\cite{wisth2022vilens} 
combines IMU, kinematics, LiDAR, and camera data using factor graphs to ensure reliable estimation, even when individual sensors may fail. 
Nevertheless, it is important to note that VILENS has not demonstrated real-time feedback control, and thus remains untested in fully closed-loop operations.
More recently, Leg-KILO~\cite{ou24legkilo} combines LiDAR odometry with kinematic and inertial measurements to estimate the robot's pose, heavily relying on loop closure to minimize the drift. Loop closures significantly enhance state estimation by reducing drift. However, they might introduce sudden changes in the estimated trajectory. For this reason, state estimators performing loop closures are typically not used to provide feedback directly to controllers that rely on smooth, continuous feedback for real-time applications.  Additionally, in situations where loop closures are infrequently, such as in a long corridor, the system will experience significant drift over an extended period.
\looseness=-1

\subsection{Contribution and Outline}
\label{sec:contribution}   
In this letter, we introduce MUSE, an innovative state estimation framework for legged robots that builds upon our earlier research in~\cite{fink20iros}. In that prior work, we deployed a nonlinear observer for attitude estimation and derived leg odometry from a quadruped model, which was then fused using a Kalman Filter (KF). With MUSE, we extend this approach into a comprehensive state estimation pipeline, incorporating exteroceptive sensors and integrating the slip detection module we presented in~\cite{nistico22sensors}, wherein a kinematics-based strategy identifies slippage in one or more legs concurrently. This enhancement enables MUSE to deliver low-drift estimates while maintaining robustness against sensor failures and operating effectively in uneven, unstructured environments.
In this context, we make the following contributions:
\begin{itemize}
    \item Integration of a slip detection module in state estimation: to the best of our knowledge, this is the first instance of a multi-sensor state estimation pipeline featuring a module specifically designed for slip detection, crucial when walking on uneven or unstructured terrain.
    \item Real-time operation: unlike previous works (e.g. VILENS, and 
    STEP~\cite{kim2022step,wisth2022vilens}), we used MUSE to provide real-time feedback to the 
    locomotion controller during an experiment conducted on the Aliengo robot.
    \item Online and offline evaluation on different platforms and scenarios: our work 
    was validated on the Aliengo robot in indoor environments on difficult scenarios,
    and on the ANYmal B300 robot, in the Fire Service College (FSC)    Dataset~\cite{wisth2022vilens} (\Cref{fig:robot_frames}). 
    We demonstrate improvements of 67.6\% and 26.67\% in translational errors compared to two state-of-the-art algorithms for quadruped robots, Pronto and VILENS respectively, along with a 45.9\% reduction in absolute trajectory error compared to TSIF. Additionally, MUSE shows superior performance in both rotational error and frequency compared to the LiDAR-inertial odometry system, DLIO. 
\end{itemize}
For the benefit of the community, we released MUSE's code under an open-source license. 
The code is available at \textbf{\url{https://github.com/iit-DLSLab/muse}}.
The remainder of this article is presented as follows:
\Cref{sec:implementation} describes the formulation of the proposed state estimator; 
\Cref{sec:experiments} presents 
the experimental results; \Cref{sec:conclusion} concludes with
final remarks.

\section{State Estimator Formulation} 
\label{sec:implementation}

\begin{figure}[t]
    \centering
    \begin{subfigure}[b]{0.26\textwidth}
        \includegraphics[width=\textwidth]{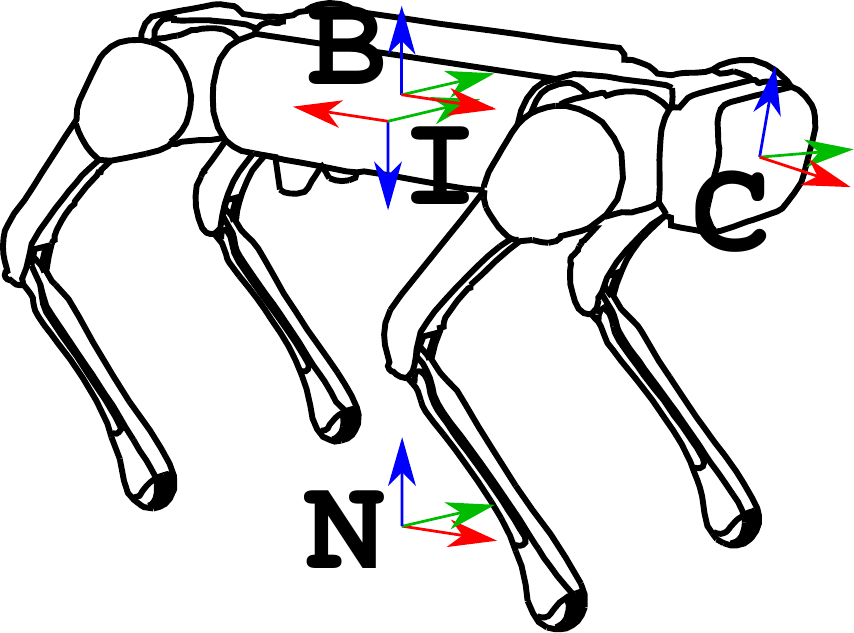}
        \caption{Aliengo frames}
        \label{fig:aliengo_frames}
        \vspace{0.3cm}
    \end{subfigure}
    \hfill
    \begin{subfigure}[b]{0.21\textwidth}
        \includegraphics[width=\textwidth]{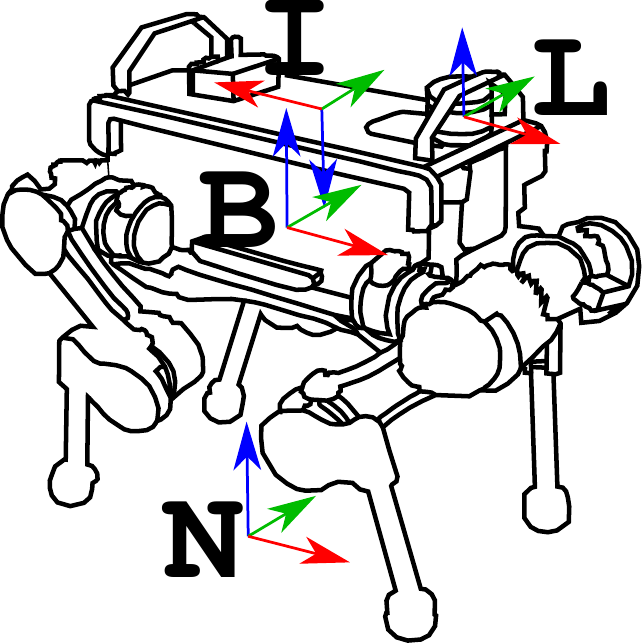}
        \caption{ANYmal B300 frames. \\
        (adapted from~\cite{wisth2021unified})}
        \label{fig:anymalbframes}
    \end{subfigure}
    \caption{Robot Reference Frames: 
    the navigation frame $\mathcal{N}$, the body frame $\mathcal{B}$, 
    the IMU sensor frame $\mathcal{I}$, the camera frame 
    $\mathcal{C}$, and the LiDAR frame $\mathcal{L}$ for ANYmal.}
    \label{fig:robot_frames}
\end{figure}

\begin{figure}
    \centering
    \includegraphics[trim={4 4 4 4},clip,width=0.48\textwidth]{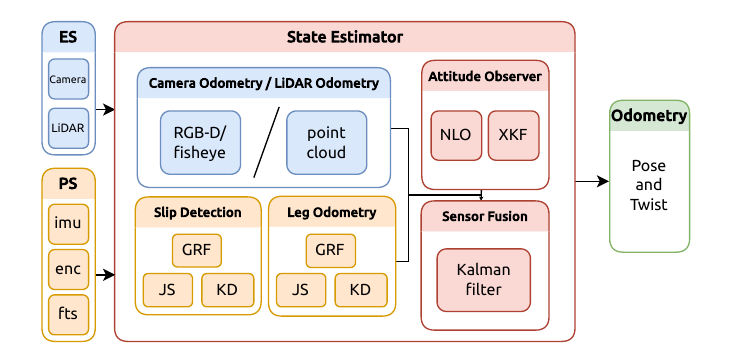}
    \caption{MUSE utilizes two exteroceptive sensors~(ES): Camera and LiDAR, and 
    three proprioceptive sensors~(PS): 
    IMU, encoders, force/torque sensors.  It comprises five main components: camera odometry or LiDAR odometry, an attitude observer 
    (AO), slip detection (SD), leg odometry (LO), and a sensor fusion algorithm (SF). The AO includes a 
    nonlinear observer~(NLO) and an eXogenous Kalman Filter~(XKF). The SD and LO include joint state (JS), 
    robot kinematics/dynamics (KD), ground reaction forces (GRF), and leg odometry models. 
    The SF utilizes a Kalman filter to estimate odometry.}
    \label{fig:se_structure}
    \vspace{-0.1cm}
\end{figure}

Our objective is to estimate the pose and twist with respect to an arbitrary
inertial navigation frame, of a quadruped robot equipped 
with a combination of proprioceptive and exteroceptive sensors, including IMUs, 
force sensors, joint sensors (encoders and torque sensors), cameras, and LiDARs. 
The state estimator consists of five major components, as shown in \Cref{fig:se_structure}: 
an exteroceptive (camera or LiDAR) odometry module, 
an attitude observer, slip detection, leg odometry, and a sensor fusion algorithm.

In this work, we used the dynamic and kinematic models of Aliengo and ANYmal~B300. However, the state estimator modules are general and can be applied to any 
legged robot with the proper sensors. 
The following reference frames (\Cref{fig:robot_frames}) are introduced: 
the navigation frame $\mathcal{N}$, which is assumed inertial,
the body frame $\mathcal{B}$ which is located at the geometric center of the trunk, 
the IMU sensor frame $\mathcal{I}$, which is located at the origin of the accelerometer of 
the IMU mounted onto the trunk of the robot, the camera frame $\mathcal{C}$ for Aliengo 
(\Cref{fig:aliengo_frames}), located at the optical center of the camera mounted in 
the front of the robot, and the LiDAR frame $\mathcal{L}$ for ANYmal 
(\Cref{fig:anymalbframes}), located at the center of the sensor mounted on top of the 
robot. The basis of the body frame is orientated forward, left, and up. We denote the reference frame of a variable using a right superscript, 
i.e., $x^n$, $x^b$, $x^i$, $x^c$, and $x^l$ denote $x$ in $\mathcal{N}$, $\mathcal{B}$, $\mathcal{I}$, $\mathcal{C}$, and $\mathcal{L}$ respectively.

The robots are equipped with a six-axis IMU on the trunk, and every joint contains an absolute encoder.
Aliengo has cameras at the front, while 
ANYmal is additionally equipped with torque sensors and a LiDAR. 
The sensors measure $\Tilde{x} = x+b_x+n_x$, where $b_x$, and $n_x$ are the bias and noise of $x$,
respectively. All of the biases are assumed to be constant or slowly time-varying, 
and all noise variables have a Gaussian distribution with zero mean.
The sensors are described as follows:
\looseness=-1
    1)~Camera: for the indoor lab experiment on Aliengo, we used a 
    lightweight tracking camera, specifically the Intel Realsense T265. It features an IMU, two fisheye lenses with 163$^\circ$ of field of view,
and the capability to provide camera odometry at up to 200 Hz.
    2)~LiDAR: for the FSC-Dataset on ANYmal B300, we used only the Velodyne VLP16 LiDAR 
    as an external sensor, whose frequency is approximately 10 Hz.
    3)~IMU: the IMU consists of a~3-DoF gyroscope and 3-DoF accelerometer.
    The accelerometer measures a specific force $f^i_i = a^i + g^i \in \mathbb{R}^3$:   
    where $a^i \in \mathbb{R}^3$ is the acceleration of the body in $\mathcal{I}$ 
    and $g^i \in \mathbb{R}^3$ is the acceleration due to gravity in $\mathcal{I}$.
    The gyroscope measures angular velocity~$\omega^i \in \mathbb{R}^3$ in  
    $\mathcal{I}$.
    4)~Encoders and Torque sensors: the absolute encoders are used to measure the joint position $q_i \in \mathbb{R}$ and joint speed $\dot{q_i} \in \mathbb{R}$, respectively. ANYmal is equipped with torque sensors that directly measure $\tau_i \in \mathbb{R}^3$, while for Aliengo, the joint torque is estimated based on the motor current.

\subsection{Camera Odometry and LiDAR Odometry}
\label{sec:cam_lid_odom}
In the lab experiments with Aliengo, odometry data is obtained using a T265 tracking camera. Conversely, for the FSC-Dataset, LiDAR odometry is employed using the KISS-ICP algorithm~\cite{Vizzo_2023}.

\subsection{Contact Estimation}
To estimate the foot contact with the ground, the contact point is assumed 
to be on a fixed point at the center of the foot. 
The contact state~$\alpha\in \mathbb{R}^{4}$ is estimated by computing the ground
reaction forces (GRFs) from the dynamics equation of motion: 
\looseness=-1

\begin{equation}
M(\bar{x})\ddot{\bar{x}}+h(\bar{x},\dot{\bar{x}}) = \bar{\tau} + J^TF_\text{grf}
\label{eq:dynamics}
\end{equation}
\noindent
where $\bar{x} = [x^T \eta^T q^T]^T \in \mathbb{R}^{18}$ is the generalized robot state, 
given by the position and attitude of the base, and the joint angles. Then
$\dot{\bar{x}} \in \mathbb{R}^{18}$ and $\ddot{\bar{x}} \in \mathbb{R}^{18}$ are the 
corresponding generalized velocities and accelerations, 
$M \in \mathbb{R}^{18\times18}$ is the joint-space inertia matrix, $h \in \mathbb{R}^{18}$ 
is the vector of Coriolis, centrifugal, and gravity forces,
$\bar{\tau}=[\underline{0} ~\tau] \in \mathbb{R}^{18}$~where~$\tau~\in~\mathbb{R}^{12}$~is 
the vector of joint torques, and finally $F_\mathrm{grf} \in \mathbb{R}^{12}$ is the vector 
of GRFs, while $J \in \mathbb{R}^{18\times12}$ is the floating base Jacobian. 
\looseness=-1

Then, assuming that all of the
external forces are exerted on the feet during the \textit{stance} phase, we first estimate the 
GRFs. Subsequently, the contact state~$\alpha_i$ for every leg~$i$ is:
\begin{equation}
	\label{eq:contact_status}
	\alpha_i = 
	\begin{cases}
		1  & \lVert F_{\text{grf},i} \rVert > F_\text{min} \\
		0  & \text{otherwise}
	\end{cases}
\end{equation}
\noindent
where~$F_\text{min} \in \mathbb{R}$ is the threshold value, and $F_{\text{grf},i} \in \mathbb{R}^{3}$ is 
the GRF of the leg~$i$. 

\subsection{Leg Odometry and Slip Detection}
Leg odometry estimates the incremental motion of the floating base 
from the forward kinematics of the legs in stable contact with the ground. 
This measurement can be formulated as either a relative pose or a velocity 
measurement. In our system, we formulate linear velocity measurements.
If there is no slippage, then the contribution of each leg 
 $\ell \in \mathbb{L}$ to the overall velocity of the base is:
 \begin{equation}
    x_{\ell}^b = -\alpha_{\ell} (J_{\ell}(q_{\ell})\dot{q}-\omega^b \times x_{\ell}^b) 
    \label{eq:leg_odom}
 \end{equation}
 and the base velocity is:
 \begin{equation}
     \dot{x}^b = \frac{1}{n_s}  \sum_{{\ell}}^{\mathbb{L}} \dot{x}_{\ell}^b
     \label{eq:base_lo}
 \end{equation}
 where $n_s =  \sum_{\ell}^{\mathbb{L}} \alpha_{\ell} $ is the number of stance legs. 
 
Leg odometry is prone to drift when the robot is walking on slippery ground. 
For this reason, we used a slip-detection algorithm, presented in our previous work
\cite{nistico22sensors}, to compensate for this characteristic drift. 
The approach is based on kinematics and makes use of velocity and 
position measurements at the ground contacts. 
We define: 
\begin{equation}
    \overline{\Delta V} = \sqrt{\sum_{i=x,y,z} \left( \frac{_d \dot{x}_{f_i}^b - \dot{x}_{f_i}^b}{|_d \dot{x}_{f_i}^b| + m} \right)^2}
    \quad \text{,} \quad
    \Delta P = \left\| _d x_{f_i}^b \right\| - \left\| x_{f_i}^b \right\|
    \label{eq:slip}
\end{equation}
\noindent
where $_d \dot{x}_f^b$ and $\dot{x}_f^b$ are the desired and measured linear velocities of the foot in $\mathcal{B}$,
while $_d x_{f_i}^b$ and $x_{f_i}^b$ are the desired and measured positions of the foot in $\mathcal{B}$, 
and $m$ is a tunable parameter used to avoid division by zero.
If $\overline{\Delta V}$ and $\Delta P$ overcome their respective 
thresholds $\epsilon_v$ and $\epsilon_p$ then a slip is detected.
We use the flag $\beta_i \in [0,1]$ for 
each leg $i$, whose value is set to $1$ if there is a slip detection, 
$0$ otherwise.

Once a single slippage or multiple slippages are detected, 
we increase the leg odometry covariance~$R_1$ in (\ref{eq:covariance})
so that the error in leg odometry 
does not negatively affect our base pose/velocity estimates.
\looseness=-1

\subsection{Attitude Observers}
\label{subsec:attitude}
To estimate the attitude, we implemented a cascaded structure composed of a 
nonlinear observer (NLO) and an eXogeneous Kalman Filter (XKF). 
The XKF linearizes about a globally stable exogenous signal from a NLO. 
The cascaded structure maintains the global stability properties from the NLO and the 
near-optimal properties from the KF. The proof of stability is explained in 
\cite{johansen:xkf}.

\subsubsection{Nonlinear Observer}
We use the non-linear observer in \cite{grip:ges}, an extension of the work
introduced in \cite{mahony:nlo2}, that makes use of symmetry properties for 
attitude estimation. 
The comprehensive equations have been elucidated in our previous publication \cite{fink20iros}.
\looseness=-1

\subsubsection{eXogeneous Kalman Filter}
The state of the filter is~$x = [q^T b^T]^T \in \mathbb{R}^7$ where 
$q \in \mathbb{R}^4$ is a quaternion, while~$b \in \mathbb{R}^3$ is the IMU's bias. 
The quaternion is used to represent the rotation as it does not contain singularities. 
The input is~$u = \omega^b \in \mathbb{R}^3$ given by the IMU's 3-axis gyroscope,
and the dynamics of the filter is:
$$
\dot{q}^n_b = \frac{1}{2} \begin{bmatrix}
0 & -(\omega^b-b^b)^T \\
(\omega^b-b^b)^T & -S(\omega^b-b^b )
\end{bmatrix}
q^n_b
$$
$$
\dot{b}^b = 0
$$
\noindent
where $S(\cdot)$ is the skew-symmetric matrix function.
We use a multiplicative error function, to respect the quaternion norm constraint $
e_q = (q^n_b)^{-1}  \otimes \hat{q}^n_b
$, 
where $\otimes$ is quaternion multiplication~\cite{markley:attitude}.

In ideal conditions, a 3-axis accelerometer and a 3-axis magnetometer provide
the measurements (feedback) to the filter. While the magnetometer can be used to 
estimate the orientation of 
an object relative to the Earth's magnetic field, there are some 
limitations when using a magnetometer for determining the orientation of a quadruped robot. 
Magnetometers are affected by local magnetic fields, due to the presence of 
high current (electric motors) and metallic objects (buildings) 
in the environment where the robots operate. 
For these reasons, we adopted another strategy to obtain the measurement vector. 
In a standard situation, the magnetometer will give a vector to a constant north. 
We implemented a ``\textit{pseudo-north}" strategy using the exteroceptive 
sensors (LiDAR or camera) for external odometry. This outputs the rotation from the sensor 
local frame ($\mathcal{S}$) to the sensor world frame ($\mathcal{SW}$). 
We used a constant vector that is constant in~$\mathcal{N}$, rotated it by the amount given by 
the sensor orientation, and then we rotated this measurement in the body frame $\mathcal{B}$. 
The measurement vector is 
$
z = [{f_b}^T {m_b}^T]^T \in \mathbb{R}^6
$, 
where 
$
f_b = R_i^b f_n \in \mathbb{R}^3
$ 
is the acceleration given by the accelerometer rotated in $\mathcal{B}$, and 
$ m_b = R_{s}^b R_{sw}^s[1~0~0]^T \in \mathbb{R}^3 $ 
is the ``pseudo-magnetometer measure", in which
 $[1~0~0]^T$ is a constant vector in $\mathcal{N}$ pointing to a ``pseudo" North, rotated in $\mathcal{B}$.
 \looseness=-1

Finally, the equations of the XKF are:

\begin{subequations}
    \begin{equation}
        \dot{\hat{x}} = f_x + F(\hat{x}-\bar{x}) + K(z-h_x-H(\hat{x}-\bar{x}))
        \label{eq:xkfdyn}
    \end{equation}
    \begin{equation}
        \dot{P} = FP + PF^T - KHP + Q
        \label{eq:xkfcov}
    \end{equation}
    \begin{equation}
        K = PH^TP^{-1}
        \label{eq:xkfgain}
    \end{equation}
\end{subequations}
where $F=\frac{\partial f_x}{\partial x}|_{\bar{x},u}$, 
$H=\frac{\partial h_x}{\partial x}|_{\bar{x},u}$, $\bar{x} \in \mathbb{R}^n$ is the 
bounded estimate of $x$ from the globally stable NLO.

\subsection{Sensor Fusion}
\label{subsec:sensor_fusion}
The inertial measurements are fused with the leg odometry and the camera or LiDAR odometry.
Decoupling the attitude from position and linear velocity offers a key benefit: 
the resulting dynamics become linear time-varying (LTV), ensuring inherent stability properties.
In other words, the filter will not diverge within a finite timeframe.
The KF has the following dynamics:

\begin{subequations}
    \begin{equation}
        \dot{\hat{x}} = f_x + K(z-h_x)
        \label{eq:ltvkfdyn}
    \end{equation}
    \begin{equation}
        \dot{P} = FP + PF^T - KHP + Q
        \label{eq:ltvkfcov}
    \end{equation}
    \begin{equation}
        K = PH^TR^{-1}
        \label{eq:ltvkfgain}
    \end{equation}
\end{subequations}
\noindent
where the state $x = [{x^n}^T {v^n}^T]^T \in \mathbb{R}^6$ is the position and 
velocity of the base, the input $u = (R_b^n f_i^b - g^n) \in \mathbb{R}^3$ is the 
acceleration of the base, and the vector $z$ is the vector of measurements. The dimensions 
of $z$ vary depending on the measurements. In the case of indoor experiments on 
Aliengo, in which the T265 camera is used as the only external sensor, $z$ is 
dimension $9$, because the T265 has pose and twist as outputs. This means that, in this case, 
$z~=~[{R_b^n \dot{x}_{\ell}^b}^T ~ {R_b^n \dot{x}_c^b}^T ~ {R_b^n x_c^b}^T]^T~\in~\mathbb{R}^9$ 
is given by the leg odometry (base velocity: ${R_b^n \dot{x}_{\ell}^b}^T$), and the camera velocity 
and position rotated in $\mathcal{N}$: ${R_b^n \dot{x}_c^b}^T$ and ${R_b^n x_c^b}^T$. 
On the FSC Dataset, on the other hand, since KISS-ICP has only the pose as output, the vector 
of measurements is 
$z = [{R_b^n \dot{x}_l^b}^T ~ {R_b^n x_l^b}^T]^T \in \mathbb{R}^6$, where $x_l^b$ 
is the position of the LiDAR in $\mathcal{B}$.
The Kalman gain~$K$ is a matrix $\in \mathbb{R}^{6 \times 9}$ when all the measurements 
are available, or~$\in \mathbb{R}^{6 \times 6}$ when the sensor velocity is not 
available.
$P \in \mathbb{R}^{6 \times 6}$ is the covariance matrix, and $Q \in \mathbb{R}^{6 \times 6}$ 
is the process noise. The measurement noise covariance matrix is a 
diagonal block matrix, assuming that the measurements are uncorrelated:
\begin{equation}
    R = \begin{bmatrix}
        R_1 & 0_3 & 0_3 \\ 
        0_3 & R_2 & 0_3 \\
        0_3 & 0_3 & R_3
    \end{bmatrix} 
    \quad \text{or} \quad
    R = \begin{bmatrix}
        R_1 & 0_3 \\ 
        0_3 & R_2
    \end{bmatrix}
    \label{eq:covariance}
\end{equation}
\noindent
where $R_1 \in \mathbb{R}^{3 \times 3}$ is the covariance of the leg odometry, and its values
are updated in case of slippage. $R_2 \in \mathbb{R}^{3 \times 3}$ is the covariance of the exteroceptive sensor velocity measurement
(when available), and $R_3 \in \mathbb{R}^{3 \times 3}$ is the covariance of the exteroceptive sensor position measurement.
Then
\begin{equation}
    f_x = \begin{bmatrix}
        v^n \\
        u
    \end{bmatrix}
    \quad \text{and} \quad
    F = \begin{bmatrix}
        0_3 & I_3 \\
        0_3 & 0_3
    \end{bmatrix}
\end{equation}
\noindent
where $I_3\in \mathbb{R}^{3 \times 3}$ and $0_3\in~\mathbb{R}^{3 \times 3}$ are the identity matrix and null matrix, 
respectively. For the same reason previously explained, the matrix $H\in \mathbb{R}^{6\times 9}$ or $H\in \mathbb{R}^{6\times 6}$ is: 
\begin{equation}
    H = \begin{bmatrix}
        0_3 & I_3 \\ 
        0_3 & I_3 \\
        I_3 & 0_3
    \end{bmatrix} 
    \quad \text{or} \quad
    H = \begin{bmatrix}
        0_3 & I_3 \\ 
        I_3 & 0_3
    \end{bmatrix}
\end{equation}
\noindent
The final structure of the state estimator is shown in Fig. \ref{fig:se_structure}.

We emphasize that to maintain efficient computation despite the slower arrival of exteroceptive measurements, we rely on internal measurements (IMU and joint states) for attitude estimation (\Cref{subsec:attitude}) and sensor fusion (\Cref{subsec:sensor_fusion}), applying corrections only when exteroceptive data becomes available.
\looseness=-1

\section{Experimental Results}

\begin{figure}[!t]
    \centering
        \includegraphics[width=0.48\textwidth]{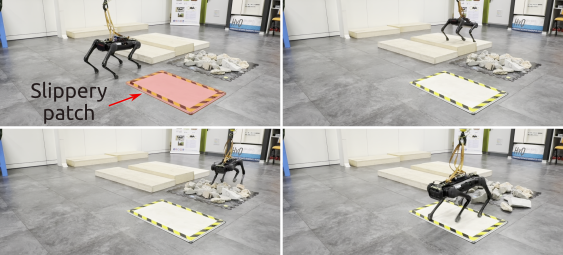}
        \caption{During the closed-loop experiment, Aliengo walked up and down the stairs, then on rocks and slippery terrain, repeating these tasks three times.}
        \label{fig:aliengo_cl}
    \end{figure}

\label{sec:experiments}
In this section, we show the results obtained on two different robotic platforms: 
Aliengo on an online lab experiment (\Cref{subsec:aliengo_online}), and
ANYmal B300 on a pre-recorded outdoor dataset (\Cref{subsec:fsc_exp}).
As our state estimator operates as an odometry system, no loop closures
(intended to recognize previously visited locations to reduce drift) have 
been executed on the estimated trajectory.

\begin{figure*}[htbp]
    \centering
    \begin{subfigure}[b]{0.48\textwidth}
        \includegraphics[width=\textwidth]{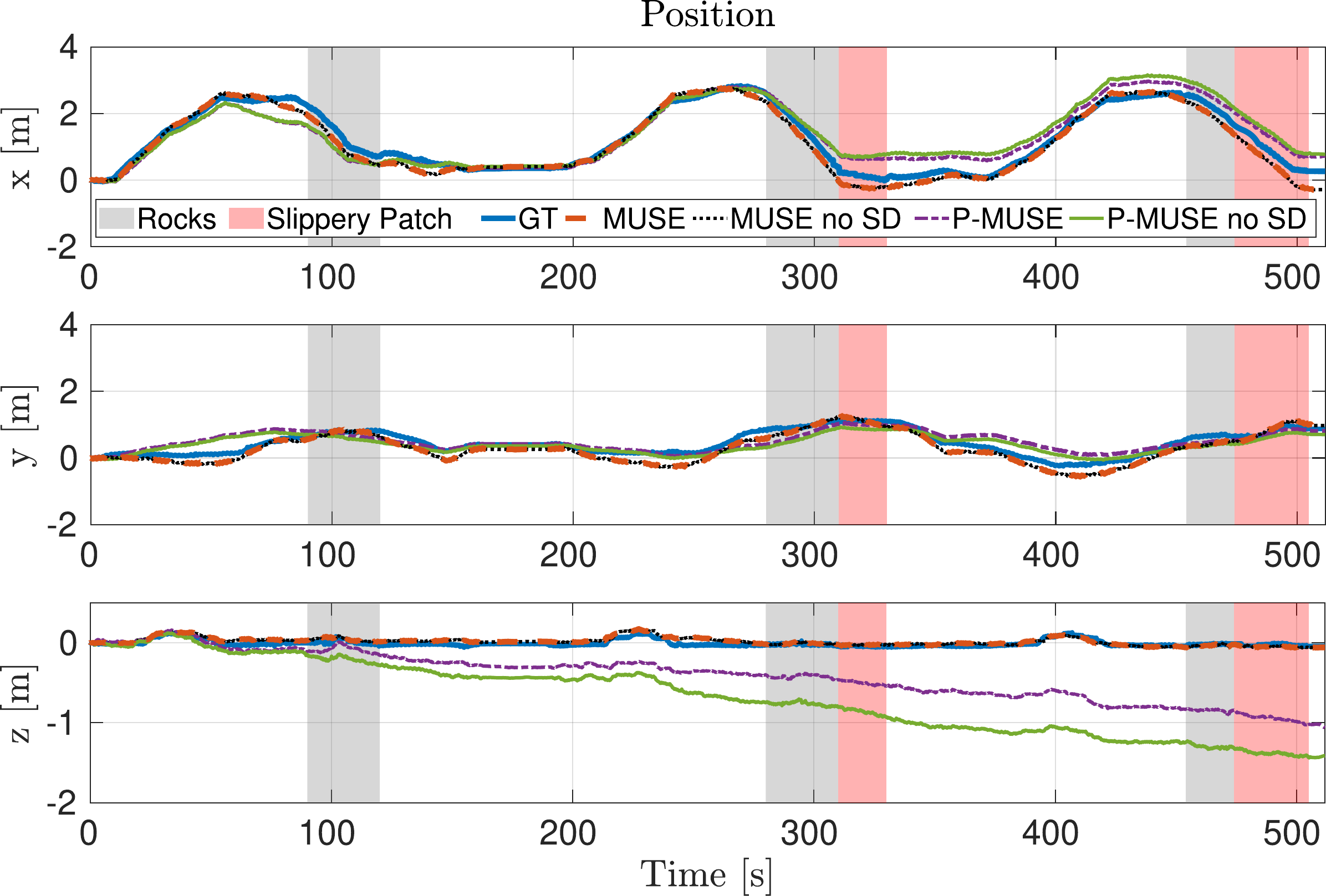}
        \caption{Ground Truth (GT) vs. estimated position.}
        \label{fig:aliengocl_pos}
    \end{subfigure}
    \hfill
    \begin{subfigure}[b]{0.48\textwidth}
        \includegraphics[width=\textwidth]{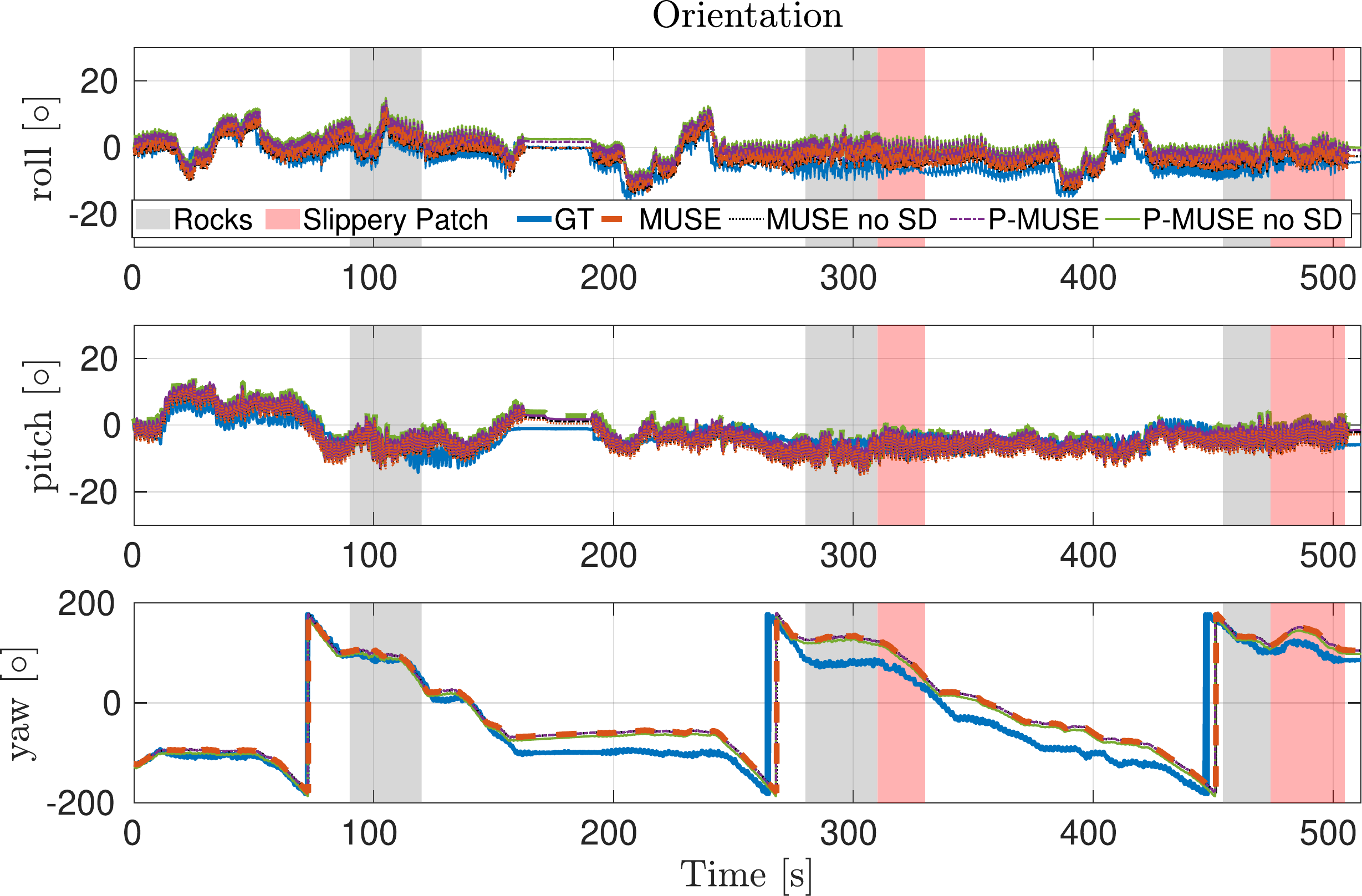}
        \caption{Ground Truth (GT) vs. estimated orientation.}
        \label{fig:aliengocl_ori}
    \end{subfigure}

    \caption{\textbf{Aliengo on uneven terrain}: Comparison of position and orientation estimations between the GT and MUSE, MUSE without the SD module (MUSE with no SD),
    Proprioceptive MUSE (P-MUSE), and P-MUSE without the 
    SD module (P-MUSE with no SD). The grey shaded areas indicate that the robot is walking on rocks, while the red ones indicate when the robot is walking on the slippery patch. The position plot (left) shows that the drift is higher when SD is not active.}
    \label{fig:comparison}
\end{figure*}
  
\begin{figure}[!t]
    \centering
        \includegraphics[width=0.48\textwidth]{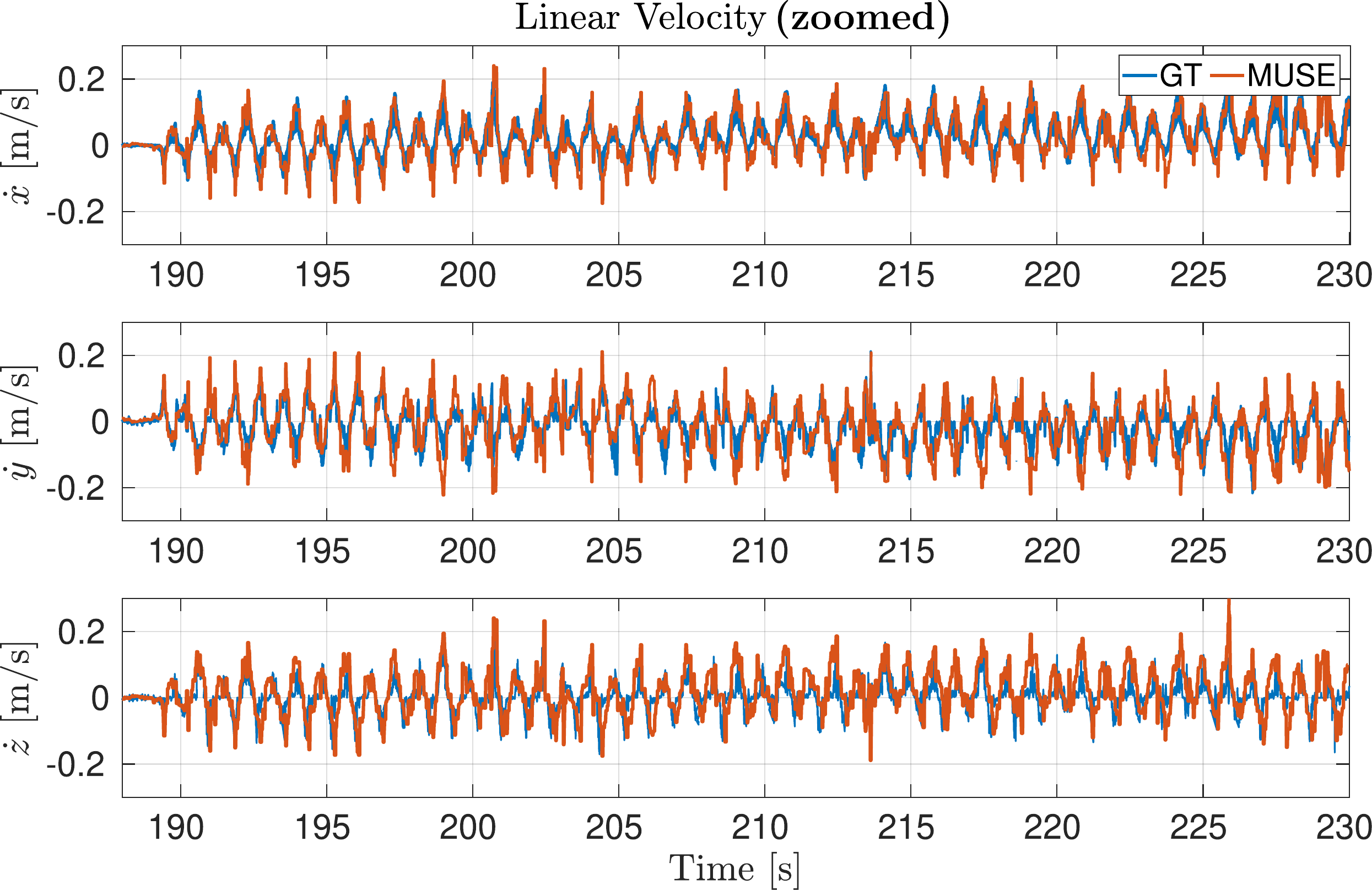}
        \caption{\textbf{Aliengo on uneven terrain}: Ground Truth (GT) vs. Linear Velocity estimated by MUSE during the closed-loop experiment, zoom into the time interval [185-230]~s.}
        \label{fig:aliengo_cl_vel}
\end{figure}

\subsection{Closing the loop with the controller}
\label{subsec:aliengo_online}
The first test is a closed-loop experiment with Aliengo. The robot walked on difficult terrain, using a crawl gait, on a trajectory of approximately 60 m, and was commanded by a joystick.
During this experiment,
the robot completed three laps around the lab, walking up and down stairs, then on rocks and slippery terrain (\Cref{fig:aliengo_cl}).

The controller is the Model Predictive Controller (MPC)
described in~\cite{amatucci2024}, that receives base pose and velocity inputs from MUSE, and gives torque commands to the joint PD controllers of the robot.
The MPC runs at~100 Hz, and the PD controller at~1000 Hz.
We ran the pipeline on an Intel NUC~i7 with~32~GB of memory. Additionally, the IMU has an acquisition frequency of~1000~Hz, as well as leg kinematics, while the camera odometry runs at~200~Hz, and MUSE runs at~1000~Hz. The average execution time of each module within MUSE is~$0.05$ milliseconds, ensuring efficient processing and real-time state updates.
  
For MUSE, we used camera orientation $R_c^n~\in~\mathbb{R}^{3\times 3} $ and IMU acceleration ($f_b~\in~\mathbb{R}^3$) as inputs for the XKF, while the linear velocity $\dot{x}_b^n~\in~\mathbb{R}^3$ from LO, linear position $x_c^n~\in~\mathbb{R}^3$ and velocity $\dot{x}_c^n~\in~\mathbb{R}^3$ from the camera, and the estimated orientation $R_b^n~\in~\mathbb{R}^{3\times3}$ from the XKF were used as inputs for SF  (\Cref{subsec:attitude,subsec:sensor_fusion}).
The ground truth was obtained using a \textit{Vicon} motion capture system.
\looseness=-1

We compared our results with those obtained using the T265’s binocular visual-inertial tracking camera, 
commonly used as a standalone state estimator in other works (for instance in~\cite{bayer2019space}). As shown 
in~\Cref{fig:aliengocl_pos,fig:aliengocl_ori}, both position and orientation estimated by MUSE closely match the ground truth. Exteroceptive measurements significantly improved the 
robot’s state estimation, particularly in correcting drift, which is common in the vertical direction (z-axis) 
when relying only on proprioception. Notably, camera odometry compensates for drift when 
walking on uneven terrain (\Cref{fig:aliengocl_pos}), making the benefits of the slip 
detection (SD) module less evident. 
However, the significance of SD is highlighted when using the proprioceptive-only variant of MUSE (P-MUSE). As shown in~\Cref{fig:aliengocl_pos}, the SD module partially compensates for position drift during slippage, which arises when leg odometry becomes unreliable. In this experiment, “slippage” encompasses not only instances when the robot traverses the designated slippery patch, but also any event in which a leg slips on a rock.

Additionally, since the MPC controller requires the robot’s linear velocity as feedback, 
\Cref{fig:aliengo_cl_vel} demonstrates that the estimated linear velocity closely aligns 
with the ground truth, enabling the robot to follow the desired trajectory.

\Cref{tab:ate_rpe_cl} presents the Absolute Trajectory Error (ATE) and Relative Pose Error 
(RPE)~\cite{zhang2018tum} statistics over 1 m. These results confirm that the MUSE pipeline provides accurate 
position estimates. The ATE is comparable to that of the T265 camera, 
but MUSE operates at a higher frequency with lower RPE in both translation and orientation. 
The advantage of the SD module is evident in P-MUSE, where the ATE is higher when SD is 
not used. Furthermore, it is important to note that the yaw angle is accurately estimated 
even using only proprioceptive sensors, owing to the globally stable Attitude Observer, 
which ensures bounded orientation error and prevents filter divergence in a finite time frame.

\begin{table}[!b]
    \begin{center}
        \caption{\textbf{Aliengo on uneven terrain}: ATE and RPE over 1~m ($\sim$ 60 m trajectory)}
        \label{tab:ate_rpe_cl}
        \begin{tabular}{c c c c c c}
            \toprule
                 & T265 & MUSE & MUSE  & P-MUSE & P-MUSE \\
                 &      &      & no SD &        &  no SD   \\
            \midrule
            \rowcolor{gray!15}
            ATE [m] & 0.24 & 0.24 & 0.25 & 0.57 & 0.67 \\
            RPE [m] & 0.10 & 0.08 & 0.09 & 0.10 & 0.12 \\
            \rowcolor{gray!15}
            RPE [$\circ$] & 0.35 & 0.25 & 0.26 & 0.27 & 0.27 \\
            Freq [Hz] & 200 & 1000 & 1000 & 1000 & 1000 \\
            \bottomrule
        \end{tabular}
        \end{center}
    \end{table}

\begin{table*}[b]
    \begin{center}
    \caption{\textbf{FSC Dataset}: ATE and RPE over 10 m ($\sim$ 240 m trajectory)}
        \label{tab:fsc_comparison}
        \begin{tabular}{c | c c c c c | c c c}
            \toprule
            & DLIO & Pronto & VILENS & MUSE & MUSE no SD & TSIF & P-MUSE & P-MUSE no SD \\
            \midrule
            \rowcolor{gray!15}
            ATE [m] & 0.14 & N.A & N.A. & 0.17 & 0.18 & 4.40 & 2.38 & 2.57 \\
            RPE [m] & 0.09 & 0.34 & 0.15 & 0.11 & 0.12 & 0.05 & 0.12 & 0.15 \\
            \rowcolor{gray!15}
            RPE [$\circ$] & 1.9 & N.A. & 1.14 & 1.78 & 1.85 & 1.96 & 1.93 & 1.96 \\
            Freq [Hz] & 100 & 400 & 400 & 400 & 400 & 400 & 400 & 400 \\
            \bottomrule
        \end{tabular}
    \end{center}
\end{table*}

\subsection{FSC Dataset with ANYmal B300}
\label{subsec:fsc_exp}

This section shows the results obtained by running MUSE on the Fire Service College Dataset~\cite{wisth2022vilens}.
The Fire Service College is a firefighting training facility located in the UK. One of the test areas represents a simulated, industrial oil rig with a total dimension of 32.5m $\times$ 42.5m.
In the experiment, ANYmal trotted at $0.3$ m/s,  
completing three loops before returning to the initial position, 
for a total of $240$ m distance covered in 33~min. The environment was challenging 
due to the presence of standing water, oil residue, gravel, and mud. 
For this experiment, we show results using LiDAR as an exteroceptive sensor.
The orientation from LiDAR odometry was 
used as an external measurement in the attitude estimate (Sec. \ref{subsec:attitude}), 
and the position estimate was used as a measurement in the sensor fusion KF (Sec. \ref{subsec:sensor_fusion}). 
 
The ground truth (GT) trajectory was obtained with millimetric accuracy by combining the 
absolute sparse positions taken from a \textit{Leica Total Station T16}, 
and a SLAM system based on ICP registration and IMU.

We computed the ATE and RPE over 10 m,
on the entire trajectory ($240$~m). 
The performance in terms of ATE and RPE was benchmarked against other state-of-the-art 
state estimators: DLIO~\cite{chen23dlio}, a LiDAR-inertial odometry algorithm, 
and three state estimators tailored for quadruped robots,
Pronto~\cite{pronto2020camurri}, VILENS~\cite{wisth2022vilens} and TSIF~\cite{bloesch2018twostate}.
Pronto and VILENS fuse exteroceptive and proprioceptive measurements, while TSIF 
uses only proprioceptive data. All of these are odometry systems that do not utilize 
loop closures, with results shown in~\Cref{tab:fsc_comparison}.

Compared to DLIO, MUSE achieved a similar ATE and translational RPE, with a difference of 
only 3 cm and 2 cm, respectively. However, MUSE showed a lower rotational RPE and operated 
at a higher frequency. Specifically for this experiment, MUSE runs at~400~Hz, because leg kinematics and IMU run at~400~Hz, whereas DLIO operates at 100 Hz on average. While incorporating leg kinematics 
introduces slightly higher ATE due to noisy leg odometry, it makes the estimator more robust 
and faster in terms of frequency. Importantly, fusing different sensor modalities helps compensate 
for the limitations of individual sensors. Fusion does not always produce a more accurate estimate 
than using a single sensor, but it provides more robust estimation by allowing each sensor 
to compensate for potential failures of others. Furthermore, sensor fusion enables the estimator 
to reach higher frequencies by relying on high-frequency inputs. 

\begin{figure}[tbp]
    \centering
    \begin{subfigure}[b]{0.48\textwidth}
        \includegraphics[width=\textwidth]{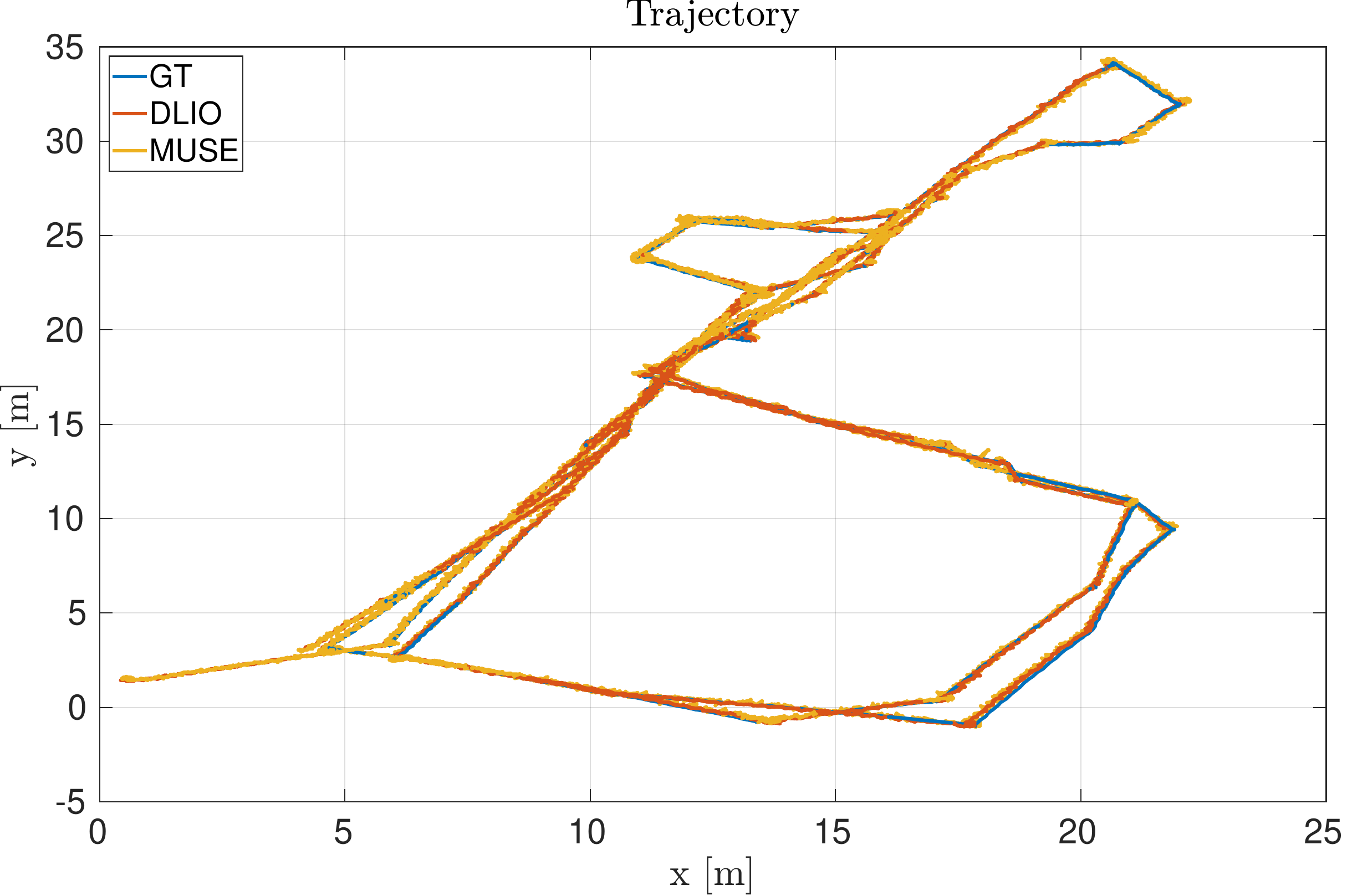}
        \caption{FSC-Dataset: ground-truth trajectory (blue) vs. estimated trajectories using DLIO and MUSE}
        \label{fig:gt_vs_est_extero}
    \end{subfigure}
    \hfill
    \begin{subfigure}[b]{0.48\textwidth}
        \includegraphics[width=\textwidth]{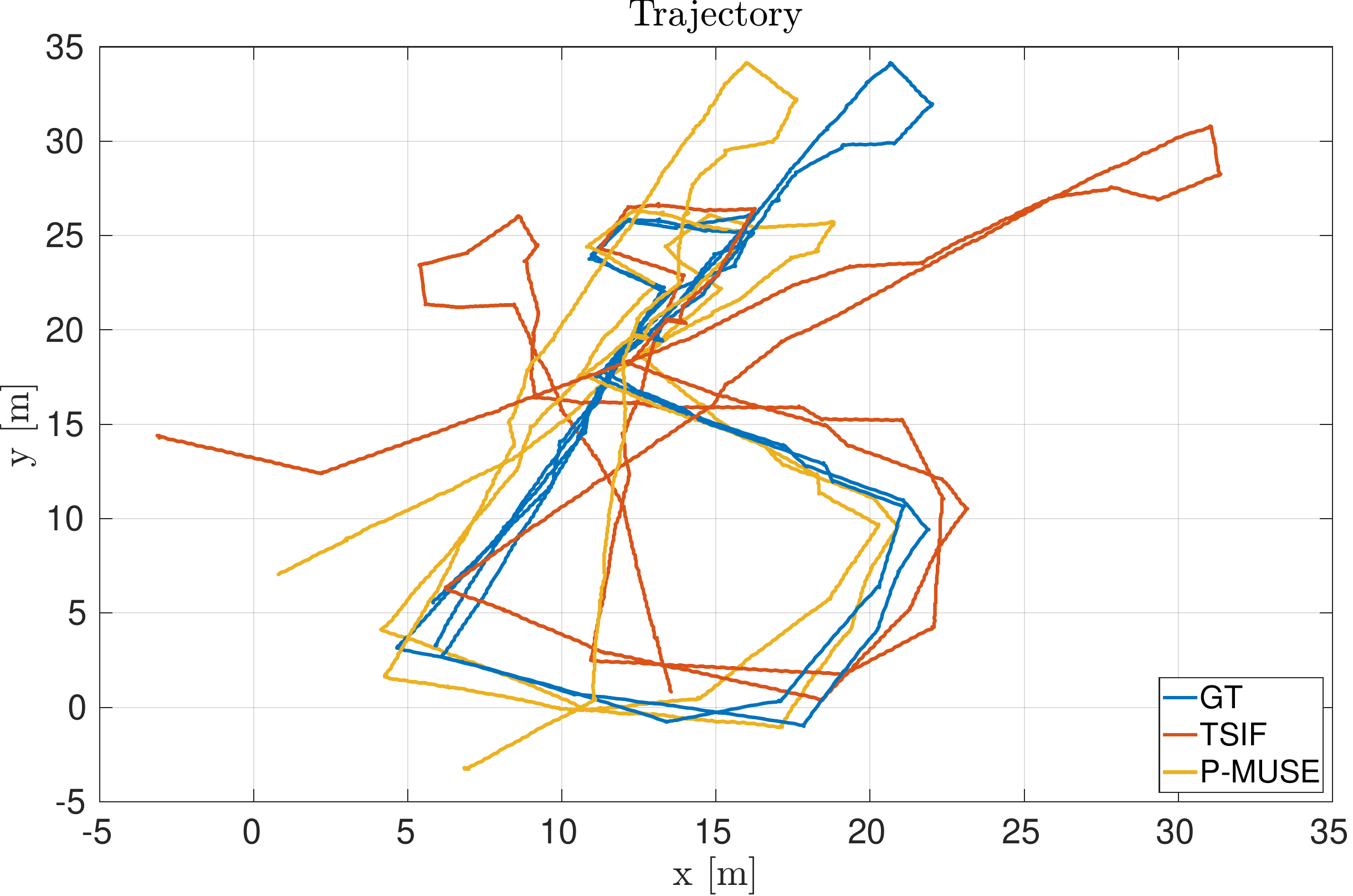}
        \caption{FSC-Dataset: ground-truth trajectory (blue) vs. estimated trajectory using TSIF and P-MUSE}
        \label{fig:gt_vs_est_proprio}
    \end{subfigure}

    \caption{\textbf{Trajectory of the FSC Dataset}: Comparison of the trajectory estimated using MUSE, P-MUSE,
    and two state-of-the-art state estimators: DLIO and TSIF.}
    \label{fig:fsc_trajectory}
\end{figure}

In comparison with Pronto and VILENS, MUSE is more accurate in terms of translational RPE, 
with improvements of 67.6\% and 26.7\%, respectively. The rotational RPE is similar to VILENS, 
although Pronto paper~\cite{pronto2020camurri} does not provide this metric, nor does either system provide ATE data. 
When comparing proprioceptive-only state estimators, P-MUSE and TSIF, our algorithm proves 
to be more accurate in terms of ATE, reducing the mean error by nearly 50\%. 
This is the most significant metric for evaluating overall trajectory discrepancy, 
reflecting global accuracy. The rotational RPE is similar between P-MUSE and TSIF, 
but P-MUSE shows a slightly higher translational RPE. This indicates that while TSIF captures 
short-term movements more precisely, small errors accumulate over time, resulting in 
inferior global accuracy compared to P-MUSE.
\Cref{fig:gt_vs_est_extero,fig:gt_vs_est_proprio} provide visual comparisons between the ground 
truth and estimated trajectories. In~\Cref{fig:gt_vs_est_extero}, we can see that DLIO and MUSE 
almost overlap with the ground truth, while in~\Cref{fig:gt_vs_est_proprio}, it is evident 
that our proprioceptive pipeline outperforms TSIF in terms of global accuracy.

\section{Discussion}
MUSE is a modular state estimator for legged robots that combines proprioceptive and exteroceptive sensor data to provide accurate and robust state estimation across various environments. Its modularity enables the integration of multiple sensor modalities, compensating for individual sensor limitations and improving overall estimation accuracy. MUSE’s real-time capability was demonstrated in the closed-loop experiment with the Aliengo robot, where it provided real-time feedback on linear velocity and orientation to the controller. Benchmarking on the FSC dataset further highlighted MUSE’s superior global and local accuracy compared to other legged robot state estimators, demonstrating its effectiveness.

The results in~\Cref{tab:fsc_comparison} underscore MUSE’s competitive performance among other estimators. Although DLIO may yield slightly lower ATE and RPE, MUSE’s up-to-400-Hz operation remains crucial for high-frequency controllers such as MPC.
Per well-established principles of cascaded controller design, the state estimation dynamics should converge significantly faster than the control loop to ensure nested-system stability~\cite{franklin2002feedback,focchi2013strategies}. This makes MUSE’s high bandwidth particularly valuable for legged robots, which require rapid and robust responses to dynamic changes in terrain, a necessity reinforced by the literature in~\cite{kleff2021high}. Furthermore, the bandwidth consistency of MUSE aligns with legged robot state estimators such as Pronto, VILENS, and TSIF, underscoring the importance of high-frequency operation in this domain. MUSE’s 400 Hz operation addresses these requirements by enabling quick, compliant reactions to disturbances while maintaining robust performance.
Considering the trade-off between estimator frequency and error metrics, we argue that a higher bandwidth offers more substantial benefits for control stability than marginally improved error statistics. Future research could explore the correlation between estimator error and control stability to provide further insights.

Additionally, slip detection is a core component of MUSE, particularly for operation on slippery or uneven terrain. The effectiveness of this module, validated in our previous work~\cite{nistico22sensors}, is central to the robustness of the overall pipeline. While threshold-based methods inherently involve a trade-off between false positives and false negatives, these were mitigated through parameter tuning and dynamic adjustments during locomotion. By incorporating slip detection, MUSE maintains accurate state estimation even when exteroceptive data is unavailable, offering more redundancy and robustness. This advantage is critical for dynamic and uneven terrains.

\subsection{Limitations}

Despite its strong performance, MUSE has some limitations that present opportunities for future work:
\begin{itemize}
    \item Friction in the robot’s joints occasionally affects the dynamics and reduces the accuracy of the contact estimation module, impacting both slip detection and leg odometry. Future work could address this by refining the contact estimation algorithm to better handle joint friction.
    \item While effective, the slip detection module does not capture all slippage events. Enhancing it to detect a broader range of scenarios, either through a probabilistic approach or machine learning techniques, is a promising avenue for improvement.
    \item MUSE assumes a fixed contact point at the center of the foot, which does not fully account for rolling motions or spherical foot geometries. This assumption can introduce errors in contact state estimation and ground reaction forces (GRFs). Developing methods to dynamically estimate and compensate for contact point variations could significantly improve robustness.
\end{itemize}
Addressing these limitations will enhance MUSE’s applicability and reliability, broadening its deployment to a wider range of environments.
\looseness=-1

\section{Conclusion}
\label{sec:conclusion}
This paper presented MUSE, a state estimator designed to improve accuracy and 
real-time performance in quadruped robot navigation. By integrating camera and 
LiDAR odometry with foot-slip detection, MUSE fuses data from multiple sources, 
including IMU and joint encoders, to provide reliable pose and motion estimates, 
even in complex environments.

Ablation studies conducted on the Aliengo robot, along with benchmarking against other state-of-the-art estimators 
using the FSC Dataset of ANYmal B300 platform, validate the robustness and adaptability of MUSE across different scenarios. 
The results demonstrate the estimator's capability to handle dynamic and challenging conditions effectively, 
ensuring reliable performance during locomotion and navigation.

Future work includes refining the contact estimation algorithm to address inaccuracies, enhancing the slip detection module to capture a broader range of scenarios, and developing methods to account for dynamic contact point variations. Additionally, implementing camera and LiDAR odometry modules for dynamic environments will enable MUSE to handle challenges introduced by moving objects, people, or animals.

\section*{Acknowledgment}

The authors would like to thank Professors Maurice Fallon (University of Oxford) and 
Marco Camurri (Free University of Bozen-Bolzano) for providing the FSC-Dataset.

\ifCLASSOPTIONcaptionsoff
  \newpage
\fi



%
\bibliographystyle{IEEEtran}
\bibliography{bibliography.bib}

%








\end{document}